\documentclass[table]{bmvc2k}

\title{Video-adverb retrieval with compositional adverb-action embeddings}

\addauthor{Thomas Hummel}{thomas.hummel@uni-tuebingen.de}{1}
\addauthor{Otniel-Bogdan Mercea}{otniel-bogdan.mercea@uni-tuebingen.de}{1}
\addauthor{A. Sophia Koepke}{a-sophia.koepke@uni-tuebingen.de}{1}
\addauthor{Zeynep Akata}{zeynep.akata@uni-tuebingen.de}{1,2}

\addinstitution{
 University of Tübingen\\
 Germany
}
\addinstitution{
 MPI for Intelligent Systems\\
 Germany
}

\runninghead{Hummel et al.}{Video-adverb retrieval with compositional embeddings}

\def\eg{\emph{e.g}\bmvaOneDot}

\usepackage{booktabs}
\usepackage{bm}
\usepackage{amsfonts} 
\usepackage{xspace}
\usepackage{wrapfig}
\usepackage{array}
\newcolumntype{C}[1]{>{\centering\let\newline\\\arraybackslash\hspace{0pt}}m{#1}}
\usepackage{multirow} 
\usepackage{pifont}
\usepackage{appendix}
\definecolor{bmvc_blue}{RGB}{0,26,102} 
\newcommand{\cmark}{\ding{51}}%
\newcommand{\xmark}{\ding{55}}%
\newcommand{\mypara}[1]{\noindent{\bf{#1}}}
\newcommand{\modelName}{\textsc{ReGAdA}\xspace}
\newcommand{\howto}{HowTo100M}
\newcommand{\vatex}{VATEX}
\newcommand{\msrvtt}{MSR-VTT}
\newcommand{\anet}{ActivityNet}
\newcommand{\air}{Adverbs in Recipes}
\newcommand{\accls}{AC\textsubscript{\textsc{cls}}}
\newcommand{\acreg}{AC\textsubscript{\textsc{reg}}}

\begin{document}

\maketitle

\begin{abstract}
Retrieving adverbs that describe an action in a video poses a crucial step towards fine-grained video understanding.
We propose a framework for video-to-adverb retrieval (and vice versa) that aligns video embeddings with their matching compositional adverb-action text embedding in a joint embedding space. The compositional adverb-action text embedding is learned using a residual gating mechanism, along with a novel training objective consisting of triplet losses and a regression target.
Our method achieves state-of-the-art performance on five recent benchmarks for video-adverb retrieval. Furthermore, we introduce dataset splits to benchmark video-adverb retrieval for unseen adverb-action compositions on subsets of the MSR-VTT Adverbs and ActivityNet Adverbs datasets. 
Our proposed framework outperforms all prior works for the generalisation task of retrieving adverbs from videos for unseen adverb-action compositions.
Code and dataset splits are available at \url{https://hummelth.github.io/ReGaDa/}.
\end{abstract}

\section{Introduction}

Fine-grained video understanding is concerned with the detailed analysis of video content beyond action recognition. This is relevant for improving and potentially accelerating video search and retrieval. While there has been significant progress in action retrieval and recognition in videos \cite{miech2020end,patrick2020support,tanberk2020hybrid, alhersh2021learning}, the fine-grained understanding of actions remains challenging. In particular,
it can be useful to perceive how an action is performed in order to better understand the action itself~\cite{doughty_action_2020,doughty_how_2022,moltisanti2023learning}. 
For instance, in addition to recognising the action \textit{cutting}, it is useful to understand details about the execution of an action, \eg \textit{cutting slowly}.
Specifically, we consider the bidirectional video-adverb retrieval task where we retrieve adverbs that match an action in a video and vice versa.

For bidirectional video-adverb retrieval, adverbs and action words can be combined in a compositional manner. The same adverb can describe multiple actions, such as \textit{cutting slowly} or \textit{dancing slowly}. 
The compositional nature of the adverb-action pairings can also be exploited when learning adverb-action representations.
Our proposed \modelName framework for video-adverb retrieval uses a \textbf{re}sidual \textbf{g}ating mechanism to compose \textbf{ad}verb-\textbf{a}ction (\modelName) representations for retrieval. 

At its core, our framework learns to align adverb representations and video representations in a shared embedding space using a novel training objective which consists of a direct regression loss between the adverb and video representations and triplet losses.
To obtain the adverb representation, the adverb and action are jointly embedded using a residual gating mechanism, which we adapted to the video-adverb retrieval task from \cite{vo2019composing}.
It models the composition as a transformation of the adverb embedding based on the action by using a gate and a residual mechanism. 
The gate facilitates the preservation of meaningful information from the adverb embeddings based on the adverb-action composition.
Our final composition is learned as a residual combination on top of the gated adverb embeddings. This allows our composed embeddings to be in the same ``feature space'' as the original adverb embeddings. 
Similar to previous works for this task, our model assumes knowledge of the ground-truth action class to perform video-adverb retrieval. 

The compositional adverb-action embeddings and our proposed training objective prove beneficial for the retrieval performance, specifically for the retrieval of unseen adverb-action compositions.
\modelName obtains state-of-the-art results on the five video-adverb retrieval benchmarks HowTo100M Adverbs~\cite{miech2019howto100m,doughty_action_2020}, VATEX Adverbs~\cite{wang2019vatex,doughty_how_2022}, ActivityNet Adverbs~\cite{caba2015activitynet,doughty_how_2022}, MSR-VTT Adverbs~\cite{xu2016msr,doughty_how_2022}, and Adverbs in Recipes~\cite{miech2019howto100m,moltisanti2023learning}. 
Furthermore, we propose two additional splits for benchmarking the retrieval of unseen adverb-action compositions on the ActivityNet Adverbs and MSR-VTT Adverbs datasets.

To summarise, we make the following contributions: 1) Our proposed method for video-adverb retrieval uses a text encoder based on a gated residual mechanism and a novel training objective. 
2) We evaluate \modelName on the challenging unseen video-adverb retrieval task and introduce new benchmark splits, compliant with zero-shot learning principles, for the retrieval of unseen adverb-action compositions based on the ActivityNet Adverbs and MSR-VTT Adverbs datasets. 3) Our framework outperforms prior work for both the seen and the unseen adverb-action composition retrieval tasks.

\section{Related work}
\mypara{Fine-grained action understanding in video retrieval.}
Early works for video understanding extended retrieval approaches for images to videos, by temporally aggregating frames in a video \cite{dong2018predicting, otani2016learning,torabi2016learning, xu2015jointly}. With the availability of large video-text datasets \cite{anne2017localizing,krishna2017dense,miech2019howto100m,oncescu2021queryd,wang2019vatex, xu2016msr,zhou2018towards,bain2021frozen}, different methods focused on sentence disambiguation \cite{chen2019cross, wray2019fine}, self-supervision \cite{alayrac2020self,rouditchenko2020avlnet,zhu2020actbert}, weakly supervised learning \cite{miech2020end,miech2019howto100m,patrick2020support}, multiple embedding experts~\cite{miech2018learning,liu2019use,gabeur2020multi}, or the use of large pre-trained models~\cite{lei2021less,luo2022clip4clip,wu2022cap4video,park2022exposing}. 
Video-action retrieval specifically aims at retrieving videos based on an action, \eg using a verb to describe the same~\cite{hahn2019action2vec,wray2019learning}.
Moreover, \cite{chen2020fine, wray2019fine, xu2015jointly, zhukov2019cross, ge2022bridging} use nouns in addition to verbs for video-text retrieval. In a more general setting, \cite{momeni2023verbs} recently proposed to use a large language model to generate modified captions to improve verb understanding in video-language models. Different to these methods, we focus on adverbs in the video-adverb retrieval task.

\mypara{Video-adverb retrieval.}
The video-adverb retrieval task was introduced by~\cite{doughty_action_2020} along with the HowTo100M Adverbs dataset. \cite{doughty_action_2020} learns a shared representation between videos and adverbs, modelling adverb information as learned linear transformations on action class label word embeddings, similar to~\cite{nagarajan2018attributes} for object attributes. Unlike \cite{doughty_action_2020}, we choose to utilise semantic information from adverb embeddings in addition to action embeddings for modelling adverb-action compositions. 
\cite{doughty_how_2022} extends \cite{doughty_action_2020} to the low-data regime with pseudo-labelling. 
The recently proposed~\cite{moltisanti2023learning} tackles the task either as a classification or regression problem. Its video encoder builds on \cite{doughty_action_2020} with an additional projection following the attention while keeping the text representations frozen. The classification variant is trained with a cross-entropy loss for adverb classification, while the regression variant uses a regression target describing the change an adverb induced in an action embedding. 
Different from \cite{moltisanti2023learning}, we aim at learning the adverb-action representations and the video representations in a shared embedding space. Formulating the task as an alignment problem in a shared embedding space combined with compositional adverb-action representations significantly boosts the performance for video-adverb retrieval. 

\mypara{Learning with object attributes.}
Approaches for learning object-attribute pairs from images can be broadly categorized into classification~\cite{misra2017red, nagarajan2018attributes, li2020symmetry, mancini2021_compcos, naeem2021_cge} and retrieval approaches~\cite{borth2013large,chen2014inferring,isola2015discovering,nan2019recognizing,wang2013unified,wang2010discriminative,vo2019composing}.
Our adverb-action compositions are most closely related to~\cite{vo2019composing}, which proposed a residual gating mechanism for learning compositional image-text embeddings. This mechanism proved particularly useful for retrieving images using both an image and a text query, the text describing a desired modification of the query image. We adapt a similar residual gating mechanism for learning compositional adverb-action embeddings by aligning the composition with action-focused video embeddings.  

\section{\modelName framework for video-adverb retrieval}
\begin{figure*}
    \begin{center}
    \includegraphics[width=\textwidth]{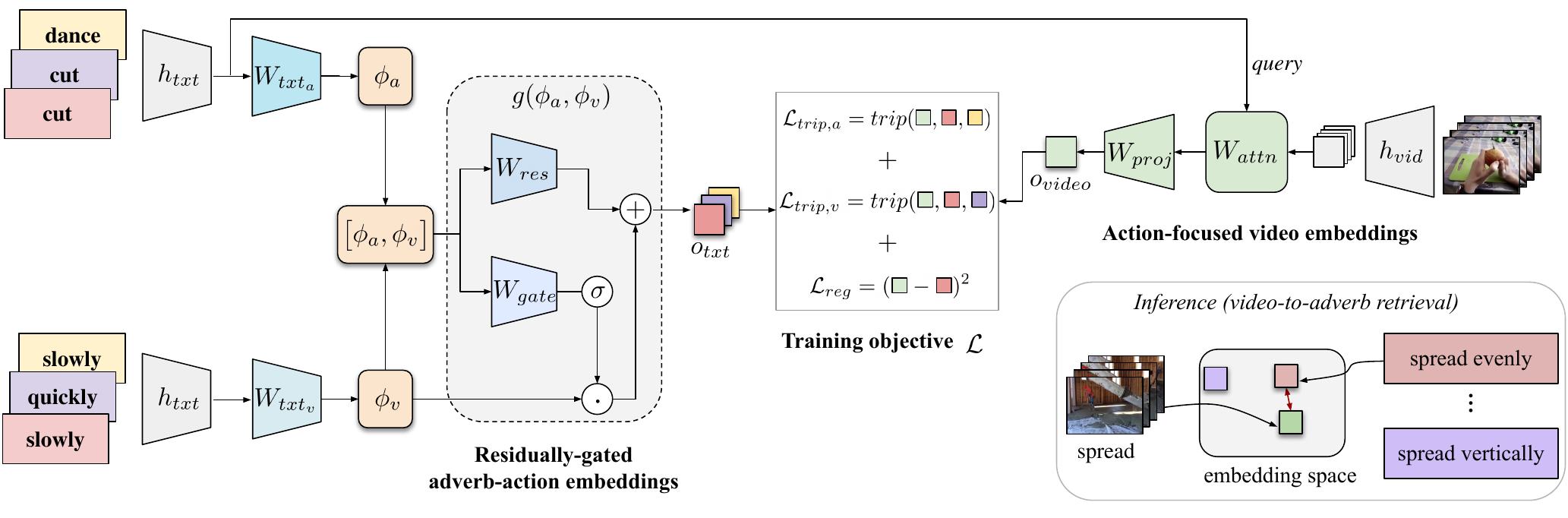} 
    \end{center}
    \caption{\textbf{Overview of our \modelName framework for video-adverb retrieval.} Our framework composes adverb-action embeddings with a gated residual between the adverbs $\phi_v$ and the concatenated action and adverb embeddings $[\phi_a,\phi_v]$. The training objective $\mathcal{L}$ aligns the learned text and video representations in a joint embedding space. For test time inference, outputs are obtained based on similarity in the embedding space. 
}
    \label{fig:model}
 \vspace{-10pt}
\end{figure*}

In this section, we provide details about our proposed \modelName framework for video-adverb retrieval which is visualised in Fig.~\ref{fig:model}. We first 
describe the video-adverb retrieval task, and then provide details about our framework. Finally, we detail our training objective and the inference procedure for retrieval.

\mypara{Task setting and dataset.} 
The adverb-to-video retrieval task aims at retrieving matching videos from a pool of videos for a given adverb. Similarly, for the video-to-adverb retrieval task, given a video, the aim is to retrieve the adverb that best describes the action depicted in the video from a pool of pre-set adverbs. 
We denote a dataset with $N$ samples, $A$ action classes and $V$ adverb classes by $\mathcal{D} = \{\mathcal{X}_{[i]}, y_{[i]}\}_{i=1}^{N}$, consisting of video data $\mathcal{X}_{[i]}$, and ground-truth action and adverb labels $y_{[i]} = \{a_{[i]}, v_{[i]}\}$ with one-hot encodings for the action $a_{[i]} \in \mathbb{R}^{{A}}$ and adverb $v_{[i]} \in \mathbb{R}^{{V}}$.
We define the sets of possible actions and adverbs as $\mathcal{A}$ and $\mathcal{V}$. The set of all possible adverb-action combinations is 
$\mathcal{C} = \mathcal{V} \times \mathcal{A}$.

Our \modelName framework learns to align video and adverb-action representations in a joint embedding space. It generates compositional textual representations for adverb-action pairs using a text encoder. Additionally, the visual information is processed in a video encoder to obtain visual representations that contain information about the adverb associated with a given action.
In the following, we describe how we obtain class label embeddings for the actions and adverbs, and how the video and text encoders process the video features and class label embeddings. 

\mypara{Residually-gated adverb-action embeddings.}
We obtain word embeddings for the action $a\in \mathcal{A}$ and for the adverb $v\in \mathcal{V}$ from a pre-trained language encoder $h_{txt}$, giving $\theta_{v}=h_{txt}(v)$, and $\theta_{a}=h_{txt}(a)$ with $\theta_a, \theta_v \in \mathbb{R}^{d_{\theta}}$.
We then apply two linear maps $W_{txt_a}, W_{txt_v}: \mathbb{R}^{d_{\theta}} \xrightarrow{} \mathbb{R}^{d_{dim}}$, such that $\phi_a=W_{{txt}_a}(\theta_a)$ and $\phi_v=W_{{txt}_v}(\theta_v)$.
The action and adverb embeddings are then further processed jointly in our text encoder. Additionally, the action word embedding $\theta_{a}$ serves as a query vector in the video encoder's attention for generating an action-focused video embedding.

Our text encoder uses a residual gating mechanism which is based on \cite{vo2019composing}. Given $\phi_{a}$ and $\phi_{v_j}$ as inputs, the output of the text encoder is defined as:
\begin{equation}
    o_{txt_j} = g(\phi_a,\phi_{v_j})=\omega_g*
    \sigma(W_{gate}(\phi_a,\phi_{v_j})) \odot \phi_{v_j}
    +\omega_r*W^{res}(\phi_{a}, \phi_{v_j}),
\end{equation}
where $j\in \{1,\cdots,V\}$, $\omega_{g}$, $\omega_{r}$ are learnable scalar weights for balancing the gating mechanism and the residual, $\odot$ is an element-wise product, and $\sigma$ the sigmoid function.
$W_{res}$ and $W_{gate}$ are modelled using MLPs with $N_r$ and $N_g$ layers respectively. For those, the input consisting of adverb and action embeddings, is first passed through a concatenation operator and batch normalisation~\cite{ioffe2015batch} is applied. The subsequent layers consist of a linear map followed by dropout~\cite{srivastava2014dropout} with probability $drop_g$ and a Leaky ReLU~\cite{xu2015empirical}.
The final layer is a linear projection to $\mathbb{R}^{d_{dim}}$.

We tackle video-adverb retrieval by aligning text and videos in a learned shared embedding space. Our residual gating mechanism models the composition as a transformation of the adverb embedding based on the action. The gating mechanism thereby allows to retain information from adverbs when actions do not provide useful semantic information.

\mypara{Action-focused video embeddings.}
A pre-trained video classification network $h_{vid}$ is used to extract a sequence of visual features $\bm{x}_{[i]} = \{x_{1}, ..., x_{t}, ..., x_{T}\}_i$, where $\bm x_{[i]}=h_{vid}(\mathcal{X}_{[i]})$ and $x_t \in \mathbb{R}^{d_{x}}$. We use $T$ to denote the number of temporal segments in a video clip.

Given a sequence of video features $\bm{x}_{[i]}$ and its associated action word embedding $\theta_{a_{[i]}}$ (for easier readability, we omit the subscripts $_{[i]}$), we obtain action-focused video embeddings using a similar mechanism as the one proposed in \cite{doughty_action_2020}. 
The video embeddings are obtained using weak action-level ground-truth in the multi-head attention mechanism~\cite{vaswani2017attention}. The action word embedding $\theta_a$ serves as the query in the attention to focus on parts of the video that are relevant to the given action, and ignore the temporal segments that may be relevant to other actions.

For the multi-head attention, we map the video features $\{x_t\}_{t \in [1,T]}$ to keys and values using linear mappings $W_k: \mathbb{R}^{d_{x}} \xrightarrow{} \mathbb{R}^{d_{head_x} H_x}$, $W_v: \mathbb{R}^{d_{x}} \xrightarrow{} \mathbb{R}^{d_{head_x} H_x}$ with $H_x$ heads and a dimension of $d_{head_x}$ per head. We also map the action word embeddings $\theta_a$ to queries with $W_q: \mathbb{R}^{d_{\theta}} \xrightarrow{} \mathbb{R}^{d_{head_x} H_x}$. 
For each attention head $j$, we have
\begin{equation}
    p^j_{attn} = g^{DL}_{attn}\left(softmax \left( \frac{W^j_q(\theta_a)^T W^j_k(\bm x)}{\sqrt{d_{head_x}}} \right)\right) W^j_v(\bm x),
\end{equation}
where $g^{DL}_{attn}$ denotes dropout with probability $drop_{attn}$.

We apply a linear mapping $W_{attn} : \mathbb{R}^{d_{head_x} H_x} \rightarrow \mathbb{R}^{d_{dim}}$ to aggregate the per-head attention giving the output video embedding $o_{attn} = W_{attn}([p^1_{attn},\cdots,p^H_{attn}])$.
The final output is obtained with an MLP, $W_{proj}: \mathbb{R}^{d_{dim}} \rightarrow \mathbb{R}^{d_{dim}}$, 
\begin{equation}
    o_{video}=W_{proj}(o_{attn}),
\end{equation}
where each of the $N_{proj}$ layers of $W_{proj}$ consists of a linear layer $W_{proj}^l: \mathbb{R}^{d_{dim}} \rightarrow \mathbb{R}^{d_{dim}} $, layer normalisation~\cite{ba2016layer} $g_{proj}^{LN}$, ReLU~\cite{nair2010rectified} $g_{proj}^{ReLU}$, and dropout $g_{proj}^{DL}$ with probability $drop_{proj}$.

\mypara{Training objectives.}
Our \modelName framework is trained with triplet losses based on \cite{doughty_action_2020} and with a direct regression loss between the video and text embeddings. 
We consider the triplet loss function $trip(a, p, n) = max(0, \|a - p\|_2 - \|a - n\|_2 +\mu)$, with the anchor embedding $a$, the embeddings for the positive and negative samples $p$ and $n$, and the margin $\mu$.
The \textbf{action triplet loss} encourages the alignment of the video representation $o_{video}$ and text embeddings with the matching action as opposed to a sampled negative action $\phi_{\bar{a}}$. For this, we use the video embedding $o_{{video}}$ as the anchor, the text embedding with ground truth action $\phi_{a}$ and adverb $\phi_{v}$ as the positive sample, and the text embedding of the same adverb but different action $\phi_{\bar{a}_i}$ as a negative:
\begin{equation}  
\mathcal{L}_{trip,a} = \frac{1}{n} \sum_{i=1}^{n} trip(o_{{video}_i},g(\phi_{a_i}, \phi_{v_i}),g(\phi_{\bar{a}_i}, \phi_{v_i})) \quad\text{for } \phi_{\bar{a}_i} \neq \phi_{a_i}.
\end{equation}
We use an \textbf{adverb triplet loss} to push text embeddings containing the adverb antonym $\phi_{\bar{v}}$ away from the ground-truth text embedding:
\begin{equation}
    \mathcal{L}_{trip,v} = \frac{1}{n} \sum_{i=1}^{n} trip(o_{{video}_i},g(\phi_{a_i}, \phi_{v_i}), g(\phi_{a_i}, \phi_{\bar{v}_i})).
\end{equation}
By restricting the negative samples for adverbs to their antonyms, the loss does not punish potential ambiguities of actions in videos (\eg a drawer being opened slowly can at the same time be opened partially but not quickly). 
Our \textbf{regression loss} directly minimises the distance between the output video and text embeddings: 
\begin{equation}
    \mathcal{L}_{reg} = \frac{1}{n} \sum_{i=1}^{n} (o_{{video}_i} - g(\phi_{a_i},\phi_{v_i}))^2.
\end{equation}
The final loss is computed as the weighted sum of the above losses according to
\begin{equation}
    \mathcal{L} = \lambda_{a} * \mathcal{L}_{trip,a} + \lambda_{v} * \mathcal{L}_{trip,v} + \lambda_{reg} * \mathcal{L}_{reg}, 
\end{equation}
with hyperparameters $\lambda_{a}$, $\lambda_{v}, \lambda_{reg} \in \mathbb{R}$.

\mypara{Retrieving adverbs and videos (inference).} 
Similar to~\cite{doughty_action_2020}, we evaluate our method on adverb-to-video and video-to-adverb retrieval given the ground-truth action $a$. For video-to-adverb retrieval, given a video $\bm x$ and action query $a$, we embed the video to obtain $o_{video}$, and we obtain embeddings for $j$ adverb-action combinations $o_{txt_j}$ for $j \in \{1,\cdots, V\}$. Using the cosine similarity metric we rank all the text embeddings $o_{txt_j}$ by their similarity to the query video embedding $o_{video}$ and we consider the highest-ranked pair as the retrieved adverb. 

For adverb-to-video retrieval, given an adverb $v$ and action $a$ that are embedded to $o_{txt}$, we define the set of test videos containing action $a$ as $\Gamma$. We rank all video embeddings $o_{video_j}$ for videos in $\Gamma$ using the similarity computed between each $o_{video_j}$ and $o_{txt}$ and select the video which is closest to $o_{txt}$. 

\section{Video-adverb retrieval benchmarks}
In this section, we provide details about the datasets used in our experiments. In particular, we use five datasets for video-adverb retrieval. Furthermore, we propose two new dataset splits for the task of retrieving adverbs from videos for unseen adverb-action compositions. 

\mypara{Video-adverb retrieval datasets.}\label{dataset_section}
HowTo100M Adverbs~\cite{doughty_action_2020} consists of 5,824 video clips with annotations for 6 adverbs and 72 actions. In the following, we refer to HowTo100M Adverbs as \textbf{HowTo100M}.
The recently proposed \textbf{\air} dataset has 10 adverbs, 48 actions and 7,003 videos. 
VATEX Adverbs~\cite{doughty_how_2022} dataset has, with 34 adverbs and 135 actions, the largest variety of annotated adverbs and actions, consisting of 14,617 videos. We refer to VATEX Adverbs as \textbf{VATEX}. ActivityNet Adverbs~\cite{doughty_how_2022} consists of 3,099 videos with 20 adverbs and 114 actions. We refer to it as \textbf{\anet}. MSR-VTT Adverbs~\cite{doughty_how_2022} is made up of 1,824 videos with 18 adverbs and 106 actions. In the following, we call this dataset \textbf{\msrvtt}.

\mypara{Unseen adverb-action compositions splits.}
We strive to explore the ability to recognise adverbs for novel adverb-action combinations.
\cite{doughty_how_2022} proposed a dataset split for unseen compositions at test time for the \vatex\ dataset. 
Using the available videos in \vatex\ from~\cite{moltisanti2023learning}, we replicate this split for the S3D video and text features used in this work, by omitting unavailable videos.
We additionally propose new splits for unseen compositions on the \anet\ and \msrvtt\ datasets. We exclude HowTo100M Adverbs and \air, as both are subsets of HowTo100M which was used for pre-training the text and S3D video model. Hence, this would not comply with zero-shot learning principles. 

To create splits for \anet\ and \msrvtt, we follow the protocol in~\cite{doughty_how_2022}: We first split the set of possible adverb-action compositions into two non-overlapping sets, so that all adverbs and all actions are present in both sets, but individual compositions are only contained in one of the sets.
We additionally constrain the compositions for each set so that for a given adverb-action composition, its antonym-action composition is assigned to the same set. 
\begin{wraptable}{r}{6.2cm}
\scriptsize
\centering 
\begin{tabular}{l|cccc}
\toprule
\textbf{Dataset} & \# tr (s) & \# t (s) & \# tr (p) & \# t (p) \\
\midrule
VATEX & 6603 & 3293 & 319 & 316   \\
MSR-VTT & 987 & 454 & 225 & 225    \\
ActivityNet & 1490 & 848 & 635 & 543  \\ 
\bottomrule
\end{tabular}
\vspace{1.2em}
\caption{Statistics of the proposed dataset splits for the retrieval of unseen adverb-action compositions on the MSR-VTT and ActivityNet datasets. (tr: train, t: test, s: video samples, p: adverb-action pairs)}	
\label{tab:dataset-statistics}
\end{wraptable} 
We assign the videos from one of the sets to the training set and split the videos of the other half into two different sets, assigning half of the instances in each composition to the test set and the other to an unlabelled set (which is used to train \cite{doughty_how_2022} with pseudo-labelling). 
Table~\ref{tab:dataset-statistics} shows details about the replicated split for \vatex, and for our proposed splits based on \anet\ and \msrvtt\ (full details are provided in the supplementary material).

\section{Experiments}
In this section, we provide details about the baselines, implementation details, and evaluation metrics used in this work. Video-adverb retrieval results on five benchmarks are presented in Section~\ref{sota_comparison}, and we provide model ablation studies in Section \ref{ablations}. In Section~\ref{unseen}, we investigate the transfer to unseen adverb-action compositions during inference.

\mypara{Baselines.}
We report results for the \textbf{Prior} and \textbf{S3D pre-trained} baselines from \cite{moltisanti2023learning}. \textbf{Prior} does not require any training but it uses the data distribution and adverb frequency for scoring. 
\textbf{S3D pre-trained} is also training-free and uses the similarity between frozen video and text representations from the S3D backbone jointly trained on video and text.
\textbf{TIRG~\cite{vo2019composing}} employs a similar residual gating mechanism as \modelName for image-text retrieval. To adapt it to the video domain, we use the same video encoder as our method. Different from \modelName, it models the composition as a transformation of the action embedding and uses a classification-based training objective.
We also compare our framework to \textbf{Action Modifier}~\cite{doughty_action_2020} and to the recently proposed \textbf{AC} frameworks~\cite{moltisanti2023learning}. AC tackles the task either as a classification (\accls) or regression (\acreg) problem. 

\mypara{Implementation details.} \label{implementation_details}
We use the video and text features provided by~\cite{moltisanti2023learning} which were extracted using a frozen S3D model that was jointly pre-trained on video-text pairs from HowTo100M~\cite{miech2019howto100m}. Here, $d_{x} = 1024$, $T$ is the length of the video in seconds, and $d_{\theta}=512$.
\modelName uses an internal embedding dimension $d_{dim}=400$.
We use $N_{g}=2$, except for \howto\ and \air\ where $N_g=3$ and $N_g=4$ respectively. Additionally, we set $N_r=2$ except for \air\ where we use $N_r=3$. The dropout probability in the residual gating mechanism is $drop_{g}=0.6$ for all datasets but \air\ and \howto\ where we use $drop_{g}=0.7$. The loss hyperparameters are chosen as $\lambda_{a}=1$ for all datasets and $\lambda_{v}=2.0$ for all datasets, except for $\lambda_{v}=1.5$ on \air. Furthermore, we use a $\lambda_{reg}=1.0$ for all dataset except for \howto\ where $\lambda_{reg}=1.5$.
We train with a batch size of $512$, and employ the Adam~\cite{kingma2014adam} optimizer with $\beta_1 = 0.9$, $\beta_2 = 0.999$, and weight decay $10^{-5}$. Our method is trained for 2000 epochs using a learning rate of $10^{-5}$ for all datasets with the exception of \howto\ where we use $3*10^{-5}$. We follow \cite{moltisanti2023learning}, and train all baselines for 1000 epochs using a learning rate of $10^{-4}$. 
We conduct all experiments on a single Nvidia 2080-Ti GPU.

\mypara{Evaluation metrics.} 
We follow~\cite{moltisanti2023learning}, and report mean Average Precision (mAP) scores for adverb-to-video-retrieval, in particular \textbf{mAP~M} (``adverb-to-video (all)'' in~\cite{doughty_action_2020}) and \textbf{mAP~W}. mAP~M is computed by ranking videos that contain the same ground-truth action according to their similarity to the adverb-action text embedding.
For mAP~W, the class scores are reweighed according to their support size in the test set. 
For video-to-adverb retrieval, we report binary antonym accuracy \textbf{Acc-A}. This is equivalent to ranking adverb-action embeddings according to their similarity to the embedded video and calculating the mAP by restricting the set of adverbs to the target adverb and its antonym (``video-to-adverb (antonym)'' in~\cite{doughty_action_2020}).
Similar to~\cite{moltisanti2023learning}, we report the best metrics independently. This means that models corresponding to each result may originate from different epochs.

\subsection{Comparison with the state of the art} \label{sota_comparison}

\begin{table}[t]
\centering
\begin{center}
    
\resizebox{\textwidth}{!}{%
\begin{tabular}{@{}l|ccc|ccc|ccc|ccc|ccc@{}}
\toprule
 & 
  \multicolumn{3}{c|}{\howto~\cite{doughty_action_2020}} &
  \multicolumn{3}{c|}{\air~\cite{moltisanti2023learning}} & 
  \multicolumn{3}{c|}{\anet~\cite{doughty_how_2022}} &
  \multicolumn{3}{c|}{\msrvtt~\cite{doughty_how_2022}} &
  \multicolumn{3}{c}{\vatex~\cite{doughty_how_2022}} \\ 
  & 
  mAP W & mAP M & Acc-A &
  mAP W & mAP M & Acc-A & 
  mAP W & mAP M & Acc-A &
  mAP W & mAP M & Acc-A &
  mAP W & mAP M & Acc-A \\ 
  \cmidrule{1-1} \cmidrule{2-7} \cmidrule{8-16}
Priors & 
  0.446 & 0.354 & 0.786 & 
  0.491 & 0.263 & 0.854 & 
  0.217 & 0.159 & 0.745 & 
  0.308 & 0.152 & 0.723 & 
  0.216 & 0.086 & 0.752 \\ 
S3D pre-tr. & 
  0.339 & 0.238 & 0.560 & 
  0.389 & 0.173 & 0.735 & 
  0.118 & 0.070 & 0.560 & 
  0.194 & 0.075 & 0.603 & 
  0.122 & 0.038 & 0.586 \\ 
TIRG~\cite{vo2019composing} & 
  0.441 & 0.476 & 0.721 & 
  0.485 & 0.228 & 0.835 & 
  0.186 & 0.111 & 0.709 & 
  0.297 & 0.113 & 0.700 & 
  0.195 & 0.065 & 0.735  \\ 
Act.~M.~\cite{doughty_action_2020} & 
  0.406 & 0.372 & 0.796 & 
  0.509 & 0.251 & 0.857 & 
  0.184 & 0.125 & 0.753 & 
  0.233 & 0.127 & 0.731 & 
  0.139 & 0.059 & 0.751 \\ 
  \accls$^\dagger$~\cite{moltisanti2023learning} & 
   0.562 & 0.420 & 0.786 & 
   0.606 & 0.289 & 0.841 &  
   0.130 & 0.096 & 0.741 & 
   0.305 & 0.131 & 0.751  & 
   0.283 & 0.108 & 0.754 \\ 
  \acreg$^\dagger$~\cite{moltisanti2023learning} & 
   0.555 & 0.423 & 0.799 & 
   0.613 & 0.244 & 0.847 &  
   0.119 & 0.079 & 0.714 & 
   0.282 & 0.114 & 0.774 & 
   0.261 & 0.086 & 0.755  \\ 
  \cmidrule{1-1} \cmidrule{2-7} \cmidrule{8-16}
  \modelName &  
  \bf{0.567} & \bf{0.528} & \bf{0.817} & 
  \bf{0.704} & \bf{0.418} & \bf{0.874} &  
  \bf{0.239} & \bf{0.175} & \bf{0.771} & 
  \bf{0.378} & \bf{0.228} & \bf{0.786} & 
  \bf{0.290} & \bf{0.113} & \bf{0.817}  \\ 
  \bottomrule
\end{tabular}
}
\end{center}

\caption{Results for adverb-to-video (mAP W/M) and video-to-adverb retrieval (Acc-A). Higher is better for all metrics. $^\dagger$ refers to updated results provided by the authors.
\vspace{-10pt}}
\label{tab:sota}
\end{table}

In Table~\ref{tab:sota}, we present adverb-to-video retrieval and video-to-adverb retrieval results with our \modelName framework on five benchmark datasets. It can be observed that \modelName outperforms the baselines across all datasets. In particular, we see more significant improvements of our framework over the prior methods for the adverb-to-video retrieval metrics (mAP W and mAP M) compared to video-to-adverb retrieval (Acc-A). 
For instance, on the \howto\ dataset \modelName outperforms \accls\ for adverb-to-video retrieval with mAP~M and mAP~W scores of 0.528 and 0.567 compared to 0.420 and 0.562. For the video-to-adverb retrieval measure Acc-A, \modelName obtains a score of 0.817 compared to 0.786 with \acreg. 

The most recent and strongest competitor \cite{moltisanti2023learning} optimises its systems using two different losses. The best results obtained from these two models are reported for each dataset and metric, showing no clear pattern as to which model variant is stronger.
Our \modelName framework consistently outperforms both model variants~\cite{moltisanti2023learning} on all metrics and datasets. We hypothesise that our framework's strong performance can be attributed to its compositional embeddings which is a key element of \modelName.
 
\subsection{Model ablations} \label{ablations}
This section analyses the impact of using different input text information, losses, and components in the text encoder on the overall video-adverb retrieval performance of \modelName.

\mypara{Input to the text encoder.}
The gating mechanism in \modelName represents the composition as a residual on top of the adverb and allows the adverb information to be retained, leveraging the action as auxiliary information.
We refer to the adverb as the \textit{main} and the action as the \textit{auxiliary} modality in \modelName.
We investigate if a compositional adverb-action word embedding $\phi_{comp}$, which directly embeds an adverb-action label pair (\eg ``\textit{cut quickly}'') with $h_{text}$, can be used as the main modality instead.
Table~\ref{tab:ablation_text_information} shows the impact of using different main and auxiliary modalities. 
\modelName obtains scores of 0.290 and 0.113 for mAP~W and mAP~M on \vatex\, compared to 0.245 and 0.078 when using $\phi_a$ as main modality and $\phi_v$ as auxiliary.
This confirms that capturing information about the adverb is crucial for solving the task.
Acc-A is less affected by the type of input information, \modelName obtains 0.817 compared to 0.806 when using $\phi_{comp}$ as main and $\phi_a$ as auxiliary modality. Overall, using $\phi_v$ as main and $\phi_a$ as auxiliary modality is most effective across datasets.

\mypara{Losses.} 
In Table~\ref{tab:loss_ablation}, we show the impact of our three loss functions, $\mathcal{L}_{trip,a}$, $\mathcal{L}_{trip,v}$, and $\mathcal{L}_{reg}$.
On \vatex, \modelName obtains a mAP~W and mAP~M of 0.290 and 0.113 compared to 0.182 and 0.074 when using only $\mathcal{L}_{reg}$. For Acc-A, \modelName obtains a score of 0.817 compared to 0.756 for $\mathcal{L}_{trip, a}+\mathcal{L}_{trip, v}$. The regression loss $\mathcal{L}_{reg}$ boosts the performance on all datasets significantly. Our novel loss combination gives the best video-adverb retrieval performance by better aligning adverb-action compositions and video representations. Previous work either only used triplet losses~\cite{doughty_how_2022, doughty_action_2020} or used a fixed textual regression target~\cite{moltisanti2023learning}. 

\begin{table}[t]
\centering
\begin{center}
    
\resizebox{\textwidth}{!}{%
\begin{tabular}{@{}cc|ccc|ccc|ccc|ccc|ccc@{}}
\toprule
 \multicolumn{2}{c|}{Text Input} & 
  \multicolumn{3}{c|}{\howto~\cite{doughty_action_2020}} &
  \multicolumn{3}{c|}{\air~\cite{moltisanti2023learning}} & 
  \multicolumn{3}{c|}{\anet~\cite{doughty_how_2022}} &
  \multicolumn{3}{c|}{\msrvtt~\cite{doughty_how_2022}} &
  \multicolumn{3}{c}{\vatex~\cite{doughty_how_2022}} \\ 
  \textit{main} & \textit{auxiliary} & 
  mAP W & mAP M & Acc-A &
  mAP W & mAP M & Acc-A & 
  mAP W & mAP M & Acc-A &
  mAP W & mAP M & Acc-A &
  mAP W & mAP M & Acc-A \\ 
  \cmidrule{1-2} \cmidrule{3-9} \cmidrule{10-17}
  $\phi_{a}$& $\phi_{v}$ & 
  0.485 & 0.390 & 0.824 & 
  0.436 & 0.221 & 0.872 &  
  0.225 & 0.147 & 0.763 & 
  0.336 & 0.144 & 0.780 & 
  0.245 & 0.078 & 0.807  \\ 
  $\phi_{comp}$& $\phi_{v}$ & 
  0.498 & 0.454 & 0.827 & 
  0.518 & 0.322 & 0.877 & 
  0.220 & 0.150 & 0.751  & 
  0.350 & 0.144 & 0.771 & 
  0.255 & 0.084 & 0.808  \\ 
  $\phi_{comp}$& $\phi_{a}$ &  
  0.503 & 0.467 & \bf 0.830 & 
  0.524 & 0.365 & \bf 0.881 &  
  0.222 & 0.147 & 0.758 & 
  0.348 & 0.146 & 0.763 & 
  0.255 & 0.090 & 0.806  \\ 
  $\phi_{v}$&  $\phi_{a}$ &  
  \bf{0.567} & \bf{0.528} & 0.817 & 
  \bf{0.704} & \bf{0.418} & 0.874 &  
  \bf{0.239} & \bf{0.175} & \bf{0.771} & 
  \bf{0.378} & \bf{0.228} & \bf{0.786} & 
  \bf{0.290} & \bf{0.113} & \bf{0.817} \\ 
  \bottomrule
\end{tabular}
}
\end{center}
\caption{Effect of using different types of input information for the text encoder in \modelName.} 
\label{tab:ablation_text_information}
\end{table}

\begin{table}[t]
\centering
\begin{center}
    
\resizebox{\textwidth}{!}{%
\begin{tabular}{@{}lll|ccc|ccc|ccc|ccc|ccc@{}}
\toprule
\multicolumn{3}{c|}{Loss} &   
\multicolumn{3}{c|}{\howto~\cite{doughty_action_2020}} & \multicolumn{3}{c|}{\air~\cite{moltisanti2023learning}} &  
\multicolumn{3}{c|}{\anet~\cite{doughty_how_2022}} & \multicolumn{3}{c|}{\msrvtt~\cite{doughty_how_2022}} & \multicolumn{3}{c}{\vatex~\cite{doughty_how_2022}}  \\ 
$\mathcal{L}_{trip,a}$ & $\mathcal{L}_{trip,v}$ & $\mathcal{L}_{reg}$  & 
mAP W & mAP M & Acc-A & 
mAP W & mAP M & Acc-A & 
mAP W & mAP M & Acc-A & 
mAP W & mAP M & Acc-A & 
mAP W & mAP M & Acc-A \\ 
\cmidrule{1-2}\cmidrule{3-4}\cmidrule{5-18}
\cmark & \xmark & \xmark &  
0.361 & 0.228 & 0.697 & 
0.429 & 0.214 & 0.836 &  
0.162 & 0.104 & 0.582 & 
0.259 & 0.138 & 0.714 & 
0.133 & 0.047 & 0.677  \\ 
\xmark & \cmark & \xmark &  
0.340 & 0.236 & 0.740 & 
0.430 & 0.213 & 0.846 &  
0.128 & 0.079 & 0.664 & 
0.260 & 0.127 & 0.737 & 
0.166 & 0.062 & 0.743  \\ 
\xmark & \xmark & \cmark &  
0.470 & 0.378 & 0.743 & 
0.468 & 0.234 & 0.839 &  
0.202 & 0.140 & 0.729 & 
0.288 & 0.186 & 0.743 & 
0.182 & 0.074 & 0.700  \\ 
\cmark & \cmark & \xmark &  
0.367 & 0.246 & 0.755 & 
0.468 & 0.239 & 0.851 &  
0.157 & 0.098 & 0.674 & 
0.273 & 0.116 & 0.737 & 
0.174 & 0.062 & 0.756  \\ 
\cmark & \cmark & \cmark &  
\bf{0.567} & \bf{0.528} & \bf{0.817} & 
\bf{0.704} & \bf{0.418} & \bf{0.874} &  
\bf{0.239} & \bf{0.175} & \bf{0.771} & 
\bf{0.378} & \bf{0.228} & \bf{0.786} & 
\bf{0.290} & \bf{0.113} & \bf{0.817} \\ 
\bottomrule
\end{tabular}
}
\end{center}
\caption{Impact of using different losses to train \modelName. For losses that are not used, the corresponding scalar weight in $\mathcal{L}$ is set to zero.
}
\label{tab:loss_ablation}
\end{table}

\mypara{Residual gating mechanism in the text encoder.} Table \ref{tab:residual_ablation} analyses the contributions of the components of the residual gating mechanism, such as the residual branch, the sigmoid, and weight sharing between the gated and residual branches. On \vatex, \modelName achieves the best results. Interestingly, sharing weights between the gated and residual branches yields only slightly weaker results, with a mAP-W score of 0.288 compared to 0.290 with \modelName. For mAP~M and Acc-A, \modelName obtains 0.113 and 0.817 compared to 0.111 and 0.815  when not using the residual. While some configurations can achieve better results in selected metrics, \modelName yields consistent state-of-the-art results across all metrics, confirming our model design choices. 

\begin{table}[t!]
\centering
\begin{center}
\resizebox{\textwidth}{!}{%
\begin{tabular}{@{}ccc|ccc|ccc|ccc|ccc|ccc@{}}
\toprule
\multicolumn{3}{c|}{Components}& 
\multicolumn{3}{c|}{\howto~\cite{doughty_action_2020}} & \multicolumn{3}{c|}{\air~\cite{moltisanti2023learning}} &
\multicolumn{3}{c|}{\anet~\cite{doughty_how_2022}} & \multicolumn{3}{c|}{\msrvtt~\cite{doughty_how_2022}} & \multicolumn{3}{c}{\vatex~\cite{doughty_how_2022}}  \\ 
R & $\sigma$ & SW & mAP W & mAP M & Acc-A & mAP W & mAP M & Acc-A & mAP W & mAP M & Acc-A & mAP W & mAP M & Acc-A & mAP W & mAP M & Acc-A \\ 
\cmidrule{1-4}\cmidrule{5-11}\cmidrule{12-18}
\cmark & \cmark & \cmark 
& 0.535 & 0.433 & 0.811 
& 0.689 & 0.404 & 0.875
& \bf{0.256} & \bf{0.190} & \bf{0.771} 
& 0.374 & 0.182 & 0.766 
& 0.288 & 0.109 & 0.808 \\ 
\cmark & \xmark & \xmark  
& 0.512 & 0.496 & 0.811 
& 0.501 & 0.269 & 0.862
& 0.234 & 0.171 & 0.770 
& 0.360 & 0.194 & 0.780  
& 0.260 & 0.098 & 0.804 \\ 
\xmark & \cmark & \xmark   
& 0.516 & 0.477 & \bf{0.817} 
& 0.562 & 0.296 & \bf{0.877}
& 0.228 & 0.169 & 0.765 
& 0.367 & 0.161 & 0.783 
& 0.283 & 0.111 & 0.815 \\ 
\cmark & \cmark & \xmark & 
\bf{0.567} & \bf{0.528} & \bf{0.817} & 
\bf{0.704} & \bf{0.418} & 0.874 &  
0.239 & 0.175 & \bf{0.771} & 
\bf{0.378} & \bf{0.228} & \bf{0.786} & 
\bf{0.290} & \bf{0.113} & \bf{0.817}  \\ 
\bottomrule
\end{tabular}%
}
\end{center}
\caption{Impact of different components in the residually-gated text encoder. R: With residual branch $W_{res}$; $\sigma$: With sigmoid; SW: Sharing weights between $W_{res}$ and $W_{gate}$. 
\vspace{-5pt}
}
\label{tab:residual_ablation}
\end{table}

\subsection{Qualitative Results}
\begin{figure*}
    \begin{center}
    \includegraphics[width=\textwidth,trim=0cm 1.37cm 0cm 0cm]{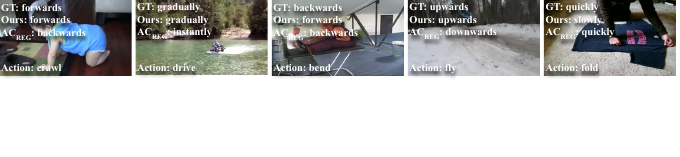} 
    \end{center}
    \caption{Example results for \modelName (Ours) on the \vatex\ dataset compared to those from \acreg. The two left examples are success cases for our model. The third and fourth example show bidirectionally performed actions that are labelled with only one of the adverbs. The right-most example shows a wrongly labelled video. Full videos are available at: \url{https://hummelth.github.io/ReGaDa}
}
    \label{fig:qualitative}
 \vspace{-10pt}
\end{figure*}

We show qualitative results for \modelName on the \vatex\ dataset in Figure~\ref{fig:qualitative}.
In particular, success cases for \modelName which \acreg\ retrieved a wrong adverb are shown below in the first and second columns. 
The third and fourth columns show videos with actions performed forwards/backwards, and upwards/downwards but labelled with only one of the adverbs. This makes both outputs plausible.
The right-most column shows an example of a wrongly labelled video for which our model retrieves the correct adverb. This confirms \modelName's strong generalisation capabilities. 
In general, we observe that \modelName better captures directional movements or speed than \acreg. It is also superior at disentangling the diverse visual effect of adverbs on different actions (\eg crawl vs.\ bend backwards). This can potentially be attributed to the compositional nature of our learned adverb-action representations.

\subsection{Generalisation to unseen adverb-action compositions}\label{unseen}

We additionally evaluate the \modelName framework on video-to-adverb retrieval for unseen adverb-action compositions, i.e.\ compositions that were not seen during training. 
We consider the existing \vatex\ benchmark and our proposed \msrvtt\ and \anet\ splits for this task (see Section~\ref{dataset_section}).
Following~\cite{doughty_how_2022}, we report binary antonym classification accuracy for video-to-adverb retrieval.
We provide additional baseline results with the CLIP~\cite{radford2021learning} model (details for this are provided in the supplementary material).
In Table~\ref{tab:unseen_compositions}, we observe that \modelName significantly outperforms \acreg\  on \vatex\ with an accuracy of 61.7 compared to 54.9. 
On \anet, \modelName obtains a score of 58.4, outperforming \cite{doughty_how_2022} with a score of 57.0. 
This is impressive given that \cite{doughty_how_2022} was additionally trained on 
\begin{wraptable}{r}{6.7cm}
\centering
\scriptsize
\begin{tabular}{l|ccc}
\toprule
\textbf{Model} & VATEX & ActivityNet & MSR-VTT \\
\midrule
CLIP~\cite{radford2021learning} & 54.5 & 55.1 & 57.0 \\
Act.~Mod.~\cite{doughty_how_2022} & 53.8  & 57.0 & 56.0\\
\accls~\cite{moltisanti2023learning} &54.3 & 55.1 & 53.7  \\
\acreg~\cite{moltisanti2023learning} &54.9 & 53.9 & 59.0  \\
\modelName & \bf 61.7 & \bf 58.4 & \bf 61.0 \\
\bottomrule
\end{tabular}
\vspace{1.2em}
\caption{Retrieval of unseen adverb-action compositions on the \vatex, \anet\ and \msrvtt\ benchmarks. \cite{doughty_how_2022} uses pseudo-labelling.}
\label{tab:unseen_compositions}
\end{wraptable}
pseudo-labelled data.
CLIP obtains an antonym accuracy of only 54.5 on \vatex, showing  a limited fine-grained retrieval capability of CLIP.
We provide a further analysis of exploiting different word embeddings for unseen compositions in the supplementary material.
Overall, our model yields better results than any prior framework for both seen (c.f.\ Table \ref{tab:sota}) and unseen compositions. 

\section{Conclusion}
In this work, we proposed a framework for video-adverb retrieval that uses a residual gating mechanism to generate compositional adverb-action representations from adverb and action word embeddings. Along with a novel training objective, our model achieves state-of-the-art results on five video-adverb retrieval benchmarks. 
Moreover, we introduce two additional dataset splits to benchmark the retrieval of unseen adverb-action compositions. Our proposed framework outperforms all prior works on this task, confirming that our text encoder results in better generalisation abilities.

\paragraph{Acknowledgements:} This work was supported by BMBF FKZ: 01IS18039A, DFG: SFB 1233 TP 17 - project number 276693517, by the ERC (853489 - DEXIM), and by EXC number 2064/1 – project number 390727645. The authors thank the International Max Planck Research School for Intelligent Systems (IMPRS-IS) for supporting T. Hummel and O.-B. Mercea. Furthermore, we would like to thank Massimiliano Mancini and Shyamgopal Karthik for helpful discussions and proofreading.

\bibliography{egbib}


\clearpage

\appendix
\vspace{6mm}
\renewcommand{\appendixpagename}{\color{bmvc_blue} \LARGE Supplementary material:\\ Video-adverb retrieval with compositional adverb-action embeddings}
\appendixpage
\addappheadtotoc
\vspace{6mm}

\section{Dataset splits for unseen adverb-action compositions}

In this section, we provide further details about our proposed dataset splits for unseen adverb-action compositions based on the  ActivityNet Adverbs~\cite{doughty_action_2020,caba2015activitynet} and MSR-VTT Adverbs~\cite{doughty_action_2020,xu2016msr} datasets. In Table~\ref{tab:supp_dataset-statistics}, we include information about the number of unlabelled samples (i.e.\ videos) and the number of unlabelled pairs (i.e.\ adverb-action compositions) in the dataset splits. 
The unlabelled samples are not used by \modelName, but we designed the splits so that we can fairly evaluate previous work~\cite{doughty_how_2022} that uses unlabelled samples for training. The number of unlabelled samples and unlabelled pairs usually ranges from 30\% to 50\% of the total number of training samples and training pairs. This is significant, as methods like~\cite{doughty_how_2022} use more training data than \modelName while performing significantly worse as observed in Table~\ref{tab:unseen_compositions} in the main paper. We refer to the ActivityNet Adverbs and MSR-VTT Adverbs datasets as ActivityNet and MSR-VTT respectively.

In addition to the ActivityNet Adverbs and MSR-VTT Adverbs datasets, we use the VATEX Adverbs dataset~\cite{wang2019vatex,doughty_action_2020}, and in particular the corresponding splits for unseen adverb-action compositions introduced in~\cite{doughty_how_2022}. However, we use the same pre-extracted features as the current state-of-the-art work~\cite{moltisanti2023learning}. As some of the videos used in the split in \cite{doughty_how_2022} are not available anymore, it is not possible to extract S3D features for those. Hence, this resulted in fewer samples in the dataset, the number of training samples being reduced from 6921 to 6603, unlabelled samples from 3469 to 3317, and test samples from 3457 to 3293. In the following, we refer to the VATEX Adverbs dataset as VATEX.

\section{Exploring the use of different word embeddings for unseen adverb-action compositions}\label{supp:sec-unseen}

Our \modelName framework composes adverb and action text embeddings in a shared embedding space. Specifically, we used a text model that was jointly trained with the S3D video model.  In this section, we show results for different choices of word embeddings. Existing and widely-adopted word embeddings like GloVe~\cite{pennington2014glove}, word2vec~\cite{mikolov2013efficient}, and fastText~\cite{bojanowski2017fasttext} rely on unsupervised learning techniques to generate vector representations of words based on their co-occurrence statistics in a large corpus of text. Specifically, word2vec and GloVe focus on co-occurrences of words, whereas fastText uses co-occurrences of n-gram characters, which can be useful when dealing with rare words.

Prior works on video-adverb retrieval leveraged GloVe embeddings of class labels~\cite{doughty_action_2020,doughty_how_2022}, while approaches in zero-shot learning commonly use word2vec or fastText embeddings as side information~\cite{xian2016latent,mancini2021_compcos,naeem2021_cge,mercea2022avca,mercea2022tcaf}. 

However, recent advances in language modelling have shown impressive progress on a variety of natural language processing tasks. For instance, large language models incorporate contextual information at the sentence level and beyond, which could result in more informative and accurate embeddings. 
To investigate their usefulness for our retrieval task, we extract word embeddings with GPT-3~\cite{gpt3} using the OpenAI API for the \texttt{text-embedding-ada-002} model. While word2vec, fastText, and GloVe provide 300-dimensional embeddings, GPT-3 embeddings have a much larger dimension of 1536. All text embeddings are projected to 400-dimensional vectors before being input into the text encoder. For CLIP~\cite{radford2021learning}, we extract visual CLIP features for each second of the video and CLIP text embeddings from the action-adverb labels (\eg \textit{cut slowly}). We then use the cosine similarity between temporally-averaged frame features and text embeddings for retrieval.

\begin{table}[t]
\centering
\begin{center}
\resizebox{\textwidth}{!}{%
\begin{tabular}{lcccccc}
\toprule
\textbf{Dataset} & \# train samples & \# unlabelled samples & \# test samples & \# pairs train & \# pairs unlabelled &\# pairs test \\
\midrule
VATEX & 6603 & 3317 &3293 & 319 & 168 & 316 \\
MSR-VTT & 987 &306 & 454 & 225 & 114 & 225 \\
ActivityNet & 1490& 634& 848& 635 & 537 & 543 \\ 
\bottomrule
\end{tabular}
}
\end{center}

\caption{Statistics of our dataset splits for the retrieval of unseen adverb-action compositions on the MSR-VTT Adverbs and ActivityNet Adverbs datasets. Statistics are also provided for the VATEX Adverbs dataset for features from~\cite{moltisanti2023learning}.
}
\label{tab:supp_dataset-statistics}
\end{table}

\begin{wraptable}{r}{.5\textwidth}
\vspace{-1.7em}
\begin{center}
\setlength{\tabcolsep}{4pt}
\resizebox{0.92\linewidth}{!}{
\begin{tabular}{l|ccc}
\toprule
Model & \vatex & \anet & \msrvtt \\
\midrule
CLIP~\cite{radford2021learning} & 54.5 & 55.1 & 57.0 \\
Act.~Mod.~\cite{doughty_how_2022} & 53.8  & 57.0 & 56.0\\
\accls~\cite{moltisanti2023learning} &54.3 & 55.1 & 53.7  \\
\acreg~\cite{moltisanti2023learning} &54.9 & 53.9 & 59.0  \\
\modelName & 61.7 & \bf 58.4 & \bf 61.0 \\
\modelName w2v & 60.5 & 53.1 & 60.0 \\
\modelName fastText & 60.8 & 53.5 & 57.3 \\
\modelName GloVe & 58.0 & 54.0 & 57.7\\
\modelName GPT-3 & \bf 63.3 & 53.5 & 60.3 \\
\bottomrule
\end{tabular}
}
\end{center}
\caption{Effect of using different types of word embeddings in our \modelName framework on the performance for retrieving unseen action-adverb compositions on the \vatex, \anet\ and \msrvtt\ benchmarks. \cite{doughty_how_2022} uses pseudo-labelling. }
\label{tab:supp_unseen_compositions}
\end{wraptable}
Table~\ref{tab:supp_unseen_compositions} shows that the choice of the text embedding results in significant performance changes, measured by the binary antonym classification accuracy. \modelName uses text embeddings jointly trained with the S3D video model like the other baselines (referred to as S3D embeddings in the following), and it is able to outperform all the baselines, as shown in the main paper. However, from Table~\ref{tab:supp_unseen_compositions} it can be observed that \modelName with S3D embeddings is outperformed by \modelName with GPT-3 embeddings on VATEX, leading to a performance of 63.3 compared to 61.7 for S3D embeddings. 
GPT-3 embeddings contain more contextual and fine-grained semantic information but suffer from a significant reduction in dimensions in the projection. We find that higher-dimensional text embeddings perform worse when training data is scarce (\eg  53.5/60.3 for GPT-3 vs.\ 58.4/61.0 for S3D on ActivityNet/MSR-VTT), likely caused by a lack of training data to learn the down-projection.
Overall, word2vec, fastText, and GloVe embeddings yield slightly worse results than S3D embeddings across datasets.

\section{Training without antonyms}
In Table~\ref{tab:results_no_antonyms}, we present the video-to-adverb and adverb-to-video retrieval performance when training without antonyms. This task was introduced in~\cite{moltisanti2023learning}. 
For the results in the main paper, \modelName is trained with antonyms as negative examples in its triplet loss. 
As it might not always be feasible to require adverb-action samples that are additionally annotated with an adverb-antonym, this scenario inspects the generalisation capabilities of \modelName to dataset settings with fewer constraints.

When training without adverb-antonyms, \modelName  randomly samples an adverb as a negative sample which is not identical to the positive adverb sample. 
As there is no access to information about the adverb-antonym during evaluation, the Acc-A metric cannot be used in this context.

In Table~\ref{tab:results_no_antonyms} we can observe that \modelName outperforms all prior methods for this task across all datasets and metrics
For example, on VATEX \modelName obtains a mAP W score of 0.292 compared to 0.283 for \accls. Moreover, \modelName obtains a mAP M score of 0.136 which significantly outperforms \accls\ with a score of 0.108. 

\begin{table*}[t]
\begin{center} 
\resizebox{\textwidth}{!}{%
\begin{tabular}{@{}l|cc|cc|cc|cc|cc@{}}
\toprule
&
\multicolumn{2}{c|}{\howto~\cite{doughty_action_2020}} &
\multicolumn{2}{c|}{\air~\cite{moltisanti2023learning}}&
\multicolumn{2}{c|}{\anet~\cite{doughty_how_2022}} &
\multicolumn{2}{c|}{\msrvtt~\cite{doughty_how_2022}} &
\multicolumn{2}{c}{\vatex~\cite{doughty_how_2022}} \\
&
mAP W & mAP M & 
mAP W & mAP M & 
mAP W & mAP M &
mAP W & mAP M & 
mAP W & mAP M \\ \midrule
Priors & 
0.446 & 0.354 &  
0.491 & 0.263 & 
0.217 & 0.159 & 
0.308 & 0.152 & 
0.216 & 0.086 \\ 
S3D pre-trained & 
0.339 & 0.238 &  
0.389 & 0.173 & 
0.118 & 0.071 & 
0.194 & 0.075 & 
0.122 & 0.038 \\ 
TIRG~\cite{vo2019composing} & 
0.441 & 0.476 & 
0.485 & 0.228 & 
0.186 & 0.111 & 
0.297 & 0.113 & 
0.195 & 0.065 \\ 
Act.~Mod.~\cite{doughty_action_2020} & 
0.408 & 0.352 &  
0.508 & 0.249 & 
0.187 & 0.127 & 
0.233 & 0.134 & 
0.144 & 0.060 \\ 
\accls$^\dagger$~\cite{moltisanti2023learning} &  
0.562 & 0.420 &  
0.606 & 0.289 & 
0.130 & 0.096 & 
0.305 & 0.131 & 
0.283 & 0.108 \\ 
\acreg$^\dagger$~\cite{moltisanti2023learning} &  
0.573 & 0.481 &  
0.667 & 0.319 & 
0.143 & 0.093 & 
0.287 & 0.121 & 
0.282 & 0.100 \\ 
\midrule
\modelName & 
\bf 0.580 & \bf 0.536 &  
\bf 0.668 & \bf 0.466 & 
\bf 0.282 & \bf 0.211 & 
\bf 0.401 & \bf 0.252 & 
\bf 0.292 & \bf 0.136 \\ 
\bottomrule
\end{tabular}
}
\end{center}
\caption{Results \textit{without} antonyms during training for adverb-to-video retrieval (mAP W/M). Higher is better for all metrics. $^\dagger$ refers to updated results provided by the authors of~\cite{moltisanti2023learning}.}
\label{tab:results_no_antonyms}
\end{table*}

\section{Comparing \modelName with CLIP}
In this section, we present additional video-adverb retrieval results with CLIP~\cite{radford2021learning} in addition to the retrieval results for unseen compositions (see Table~\ref{tab:supp_unseen_compositions}).

Similar to the experiment on unseen compositions (see Section~\ref{supp:sec-unseen}), we use the cosine similarity between temporally-averaged CLIP frame features and text embeddings for the retrieval with CLIP.
Additionally, we examine the impact of replacing the S3D video/text embeddings of \modelName with CLIP embeddings ($\modelName_{\text{CLIP}}$). 

\begin{table}[t]
\centering
\begin{center}
\resizebox{\linewidth}{!}{%
\begin{tabular}{@{}c|ccc|ccc|ccc@{}}
\toprule
& 
\multicolumn{3}{c|}{\anet} &
\multicolumn{3}{c|}{\msrvtt} &
\multicolumn{3}{c}{\vatex} \\ 
& 
mAP W & mAP M & Acc-A &
mAP W & mAP M & Acc-A &
mAP W & mAP M & Acc-A \\
\cmidrule{1-10}
S3D pre-tr. &  
0.118 & 0.070 & 0.560 &
0.194 & 0.075 & 0.603 &
0.122 & 0.038 & 0.586 \\
CLIP~\cite{radford2021learning} &
0.120 & 0.067 & 0.611 & 
0.206 & 0.084 & 0.677 & 
0.129 & 0.039 & 0.644   \\ 
$\modelName_{\text{CLIP}}$ &
0.201 & 0.151 & \bf 0.781 & 
0.352 & 0.142 & 0.784 & 
0.247 & 0.098 & \bf 0.837 \\ 
\cmidrule{1-10}
\modelName & 
\bf 0.239 &  \bf 0.175 &  0.771 & 
\bf 0.378 &  \bf 0.228 &  \bf 0.786 & 
\bf 0.290 & \bf 0.113 & 0.817  \\ 

\bottomrule
\end{tabular}%
} 
\end{center}
\caption{Comparing \modelName with CLIP as a baseline, and when replacing \modelName's S3D video/text embeddings with CLIP embeddings ($\modelName_{\text{CLIP}}$).}
 \label{tab:clip}
 \end{table}
In Table~\ref{tab:clip}, we can observe that CLIP performs marginally better than the S3D pre-trained baseline. 
Using CLIP features in \modelName improves adverb retrieval (Acc-A) slightly on ActivityNet and VATEX. However, $\modelName_{\text{CLIP}}$ is worse than \modelName for video retrieval, likely caused by inferior visual features when extracting those only from a few video frames.

\section{Seed experiments}
In Table~\ref{tab:seed_experiment}, we provide experimental results that test the robustness of our model with regard to the seeds used, as done in \cite{moltisanti2023learning}. To compute these numbers, we use four seeds and compute the mean and the standard deviation over these runs. It can be observed that \modelName achieves a higher mean than the other baselines. Furthermore, the standard deviation with our model is relatively low.

\begin{table*}[t]
\begin{center}
\resizebox{0.6\textwidth}{!}{%
\begin{tabular}{l|ccc}
\toprule
& \multicolumn{3}{c}{\air~\cite{moltisanti2023learning}} \\
& mAP W & mAP M & Acc-A \\ 
\cmidrule{1-4}
Act.~Mod.~&
$0.394 \pm 0.023$ & $0.140 \pm 0.026$ & $0.843 \pm 0.013$ \\
MLP+Act.~Mod.~& 
$0.407 \pm 0.044$ & $0.151 \pm 0.033$ & $0.842 \pm 0.012$ \\
\accls$^\dagger$ & 
$0.605 \pm 0.001$ & $0.287 \pm 0.001$ & $0.841 \pm 0.000$ \\
\acreg$^\dagger$ & 
$0.611 \pm 0.002$ & $0.239 \pm 0.007$ & $0.845 \pm 0.001$ \\
\modelName & 
$\bf {0.699} \pm \bf {0.004}$ & $\bf 0.419 \pm \bf 0.012$ & $\bf 0.876 \pm \bf 0.001$\\ 
\bottomrule
\end{tabular}
}
\end{center}
\caption{Performance of our \modelName framework on the Adverbs in Recipes dataset when using multiple random seeds. $^\dagger$ refers to updated results provided by the authors of~\cite{moltisanti2023learning}. }
\label{tab:seed_experiment}
\end{table*}

\end{document}